\documentclass{article}

\usepackage[nonatbib, preprint]{neurips_2026}

\usepackage[utf8]{inputenc} 
\usepackage[T1]{fontenc}    
\usepackage{hyperref}       
\usepackage{url}            
\usepackage{booktabs}       
\usepackage{amsfonts}       
\usepackage{nicefrac}       
\usepackage{microtype}      
\usepackage{xcolor}         
\usepackage{graphicx}
\usepackage{booktabs}

\usepackage{xspace}
\usepackage{booktabs}
\usepackage{multirow}
\usepackage{multicol}
\usepackage{xcolor,colortbl}
\usepackage{amssymb}
\usepackage{fixltx2e}
\usepackage{utfsym}
\usepackage{tablefootnote}
\usepackage{subcaption}
\usepackage{tabularx}
\usepackage{ifthen} 
\usepackage{tcolorbox}
\usepackage{amsmath}
\usepackage{cleveref}

\newcommand{\ours}{KARL\xspace}
\newcommand{\ourbench}{KVG-Bench\xspace}

\definecolor{lightergray}{HTML}{E9E9E9}
\definecolor{lighterpurple}{HTML}{E4DDF3}

\title{KARL: Knowledge-Aware Reasoning and Reinforcement Learning for Knowledge-Intensive Visual Grounding}

\author{
\textbf{Xinyu Ma}$^{1}$\thanks{Equal contribution}\enspace\textbf{,} \ 
\textbf{Ziyang Ding}$^{2}$\footnotemark[1]\enspace, \ 
\textbf{Zhicong Luo}$^{3}$, \ 
\textbf{Chi Chen}$^{4}$\thanks{Corresponding author}\enspace, \ 
\textbf{Zonghao Guo}$^{4}$, \ 
\textbf{Xuebo Liu}$^{1}$, \\
\textbf{Derek F. Wong}$^{1}$\footnotemark[2]\enspace\textbf{,} \ 
\textbf{Zhen Zhao}$^{5}$\textbf{,} \ 
\textbf{Xiaoyi Feng}$^{3}$\textbf{,} \ 
\textbf{Maosong Sun}$^{4}$ \\
\and
$^1$University of Macau, \ $^2$Shandong University,\\
$^3$Northwestern Polytechnical University, $^4$Tsinghua University, $^5$Shanghai AI Laboratory \\
{\tt\small chenchithu@gmail.com, }
{\tt\small derekfw@um.edu.mo}
}

\begin{document}

\maketitle

\begin{abstract}
Knowledge-Intensive Visual Grounding (KVG) requires models to localize objects using fine-grained, domain-specific entity names rather than generic referring expressions. 
Although Multimodal Large Language Models (MLLMs) possess rich entity knowledge and strong generic grounding capabilities, they often fail to effectively utilize such knowledge when grounding specialized concepts, revealing a knowledge–
grounding gap between internal knowledge and grounding predictions. 
To address this challenge, we propose a knowledge-aware training paradigm for KVG. 
Our approach first constructs knowledge-guided reasoning data to encourage models to activate domain-relevant entity knowledge during grounding, and then introduces KARL, a Knowledge-Aware Reinforcement Learning framework that adaptively modulates reward signals according to the model’s estimated knowledge mastery of different entities. 
To facilitate systematic evaluation, we introduce \ourbench, a benchmark spanning 10 domains with 1.3K curated test cases covering 531 images and 882 entities. 
Extensive experiments show that our approach consistently outperforms a wide range of baseline models and achieves substantially stronger cross-domain generalization on unseen categories. The data, codes, and models are released at \textcolor[RGB]{135,206,235}{https://github.com/thunlp/KARL}
\end{abstract}
\section{Introduction}
\label{sec:intro}

\begin{figure*}[t]
\centering
\includegraphics[width=1.0\linewidth]{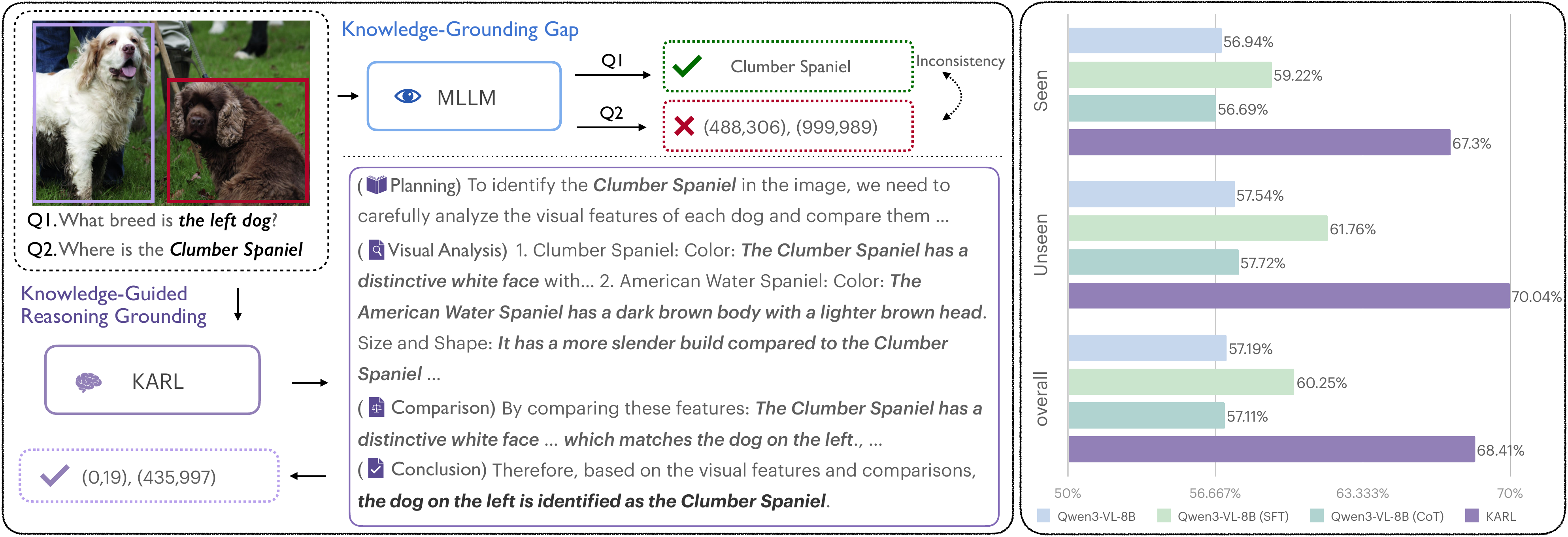}
\captionof{figure}{(a) While the MLLM can correctly recognize the entity (Q1), it fails to ground it (Q2), revealing an inconsistency between knowledge and grounding. Our method integrates knowledge-guided reasoning to bridge this gap. (b) (b) \ours achieves substantially stronger grounding performance than the baseline model and zero-shot CoT prompting, showing that knowledge-guided reasoning for KVG cannot be effectively induced by simple prompting alone.
\label{fig:teaser}} 
\end{figure*}

Human experts demonstrate fine-grained discrimination of visual concepts by flexibly leveraging domain knowledge to refine discriminative features, resulting in superior domain-specific visual perception compared to non-experts~\cite{tanaka1991object,hegde2008time, goldstein2007cognitive}. 
In contrast, recent Multimodal Large Language Models (MLLMs) exhibit rich fine-grained knowledge of diverse entities~\cite{he2025analyzing, he2026finer} and strong generic visual grounding capabilities~\cite{wang2024cogvlm, chen2024expanding, bai2025qwen3}, yet often fail to operationalize such knowledge when perception must culminate in a grounding prediction.
As illustrated in \cref{fig:teaser}a, for the same entity, a model may correctly identify the queried concept yet predict an erroneous bounding box when required to ground it in the image.
This discrepancy suggests the presence of a knowledge–grounding gap, where internal entity knowledge and grounding ability do not reliably compose into accurate knowledge-intensive grounding.

To systematically investigate this issue, we introduce \textbf{Knowledge-Intensive Visual Grounding}~(KVG), a visual grounding task that extends conventional visual grounding~\cite{kazemzadeh2014referitgame,yu2016refcoco} by requiring both fine-grained visual perception and domain-specific entity knowledge integration.
As shown in \cref{fig:teaser}a, each query in KVG utilizes domain-specific terminology~(``Clumber Spaniel'') rather than generic descriptions~(``the left dog''), and images contain visually similar distractors that necessitate knowledge-guided differentiation.
Thus, successful grounding depends not only on recognizing visual patterns, but also on aligning domain-level knowledge with spatial evidence selection.
Unlike the mathematical and geometric tasks studied by previous reasoning MLLMs~\cite{xu2024llavao1,yao2024mulberry}, KVG emphasizes the integration of entity knowledge within visual perception processes.

By requiring the composition of entity knowledge with fine-grained visual discrimination, KVG poses substantial challenges for current MLLMs.
For instance, although Qwen3-VL-8B~\cite{bai2025qwen3} achieves high accuracy on the widely used RefCOCO benchmark~\cite{yu2016refcoco}, its performance substantially declines on KVG (\cref{fig:teaser}b), indicating that strong generic grounding does not readily generalize to knowledge-intensive settings.
Moreover, even when equipped with explicit step-by-step reasoning prompting or trained with recent reasoning-guided visual grounding frameworks~\cite{bai2025univg,yu2025perceptionr}, models fail to demonstrate consistent performance improvements on KVG.
These observations suggest that the core difficulty lies not in insufficient reasoning depth, but in the ineffective alignment between fine-grained entity knowledge and grounding decisions.

To mitigate this limitation, we propose a knowledge-aware training paradigm tailored to knowledge-intensive visual grounding.
Our approach begins with constructing knowledge-guided reasoning data that encourages models to explicitly activate domain-relevant entity knowledge prior to producing grounding predictions.
By guiding models to articulate and connect entity-level knowledge with visual evidence, this stage aims to strengthen the alignment between knowledge utilization and grounding.
Building on this foundation, we introduce KARL, a Knowledge-Aware Reinforcement Learning framework that dynamically modulates optimization signals at the entity level.
Rather than applying uniform reward signals across all samples, KARL adjusts reinforcement strength according to the model’s estimated knowledge mastery of different entities, enabling differentiated optimization across heterogeneous entity types.
This design is motivated by our observation that entities vary substantially in their prior knowledge availability within MLLMs, leading to uneven grounding behavior across entities.
By explicitly coupling entity knowledge utilization with visual grounding optimization, our method aims to reduce the knowledge–grounding gap and enhance the model’s ability to leverage entity knowledge during grounding.

To facilitate systematic evaluation on KVG, we introduce \ourbench, a high-quality, expertly-curated benchmark encompassing 10 distinct domains with 1.3K manually curated test cases linked to 531 images and 882 entities.
Our extensive experiments conducted across KVG and related tasks reveal two key findings: 
(1) Knowledge-guided reasoning training consistently improves KVG performance over direct fine-tuning and generic reasoning strategies, demonstrating the effectiveness of explicitly activating entity-level knowledge during grounding; 
(2) KARL achieves substantially stronger cross-domain generalization on unseen categories, particularly compared with standard GRPO optimization, indicating that knowledge-aware reward scaling helps better generalize across entities with different knowledge levels.

In summary, our contributions are threefold:
\begin{itemize}
    \item We introduce Knowledge-Intensive Visual Grounding (KVG) task and \ourbench, a benchmark designed to evaluate models’ ability to leverage domain-specific entity knowledge for visual grounding. Our empirical observations suggest the presence of a knowledge–grounding gap in current MLLMs.

    \item We propose a knowledge-guided reasoning training strategy that constructs CoT reasoning data to encourage models to explicitly activate and align entity-level knowledge with visual evidence during grounding, differing from recent reasoning-guided grounding approaches that primarily emphasize structured reasoning depth.
    
    \item We present \ours, a Knowledge-Aware Reinforcement Learning framework that dynamically modulates optimization signals according to entity-level knowledge mastery rather than applying uniform reward schemes. This design promotes more balanced optimization across entities with heterogeneous knowledge levels and leads to improved generalization in knowledge-intensive grounding.

\end{itemize}
\section{Related Work}
\label{sec:rw}

\subsection{Multimodal Large Language Models}
\label{sec:related-mllm}

Recent years have witnessed rapid advancements in MLLMs~\cite{openai_gpt4o, liu2023visual, Liu_2024_CVPR, chen2024expanding, yao2024minicpm, bai2025qwen3}, demonstrating strong performance in tasks such as visual grounding~\cite{chen2023shikra, peng2023kosmos, you2024ferret, ma2024groma, lai2024lisa} and reasoning~\cite{shi-etal-2024-math, xu2024llavao1, deng2024r, zhang2024mathverse}.
Early MLLM-based grounding approaches such as Shikra~\cite{chen2023shikra}, Groma~\cite{ma2024groma}, and LISA~\cite{lai2024lisa} adopt end-to-end architectures that align textual queries with image regions, enabling direct prediction of bounding boxes or segmentation masks. 
While effective, these methods primarily rely on learned visual-text alignment and typically perform grounding in a single-step manner without explicit reasoning.

Inspired by recent reinforcement learning approaches for enhancing reasoning in large language models~\cite{guo2025deepseek}, recent works such as UniVG-R1~\cite{bai2025univg}, Perception-R1~\cite{yu2025perceptionr}, and Visual-RFT~\cite{liu2025visual} adopt reinforcement learning to improve visual grounding performance by strengthening the model's reasoning capabilities.
However, these approaches mainly focus on strengthening generic reasoning capabilities for grounding, without explicitly encouraging the utilization of domain-specific knowledge during the reasoning process. 
In contrast, our work introduces knowledge-guided reasoning that encourages models to connect entity-level knowledge with visual evidence during grounding.

\subsection{Visual Grounding Benchmarks}
\label{sec:related-vg}
Visual Grounding aims to locate objects or regions within an image based on a textual query~\cite{kazemzadeh2014referitgame, yu2016refcoco, qiao2020referring}.
To enable more comprehensive evaluation of MLLMs' grounding capabilities, recent studies have introduced several more challenging benchmarks~\cite{chen2020cops, chen2023advancing, lai2024lisa} in addition to the widely used RefCOCO/+/g datasets~\cite{kazemzadeh2014referitgame, mao2016generation}. 
SK-VG~\cite{chen2023advancing} explores grounding with scene-level knowledge integration, while ReasonSeg~\cite{lai2024lisa} requires models to leverage world knowledge during object localization.
In contrast, our proposed KVG task focuses on knowledge-intensive setting, requiring models to utilize fine-grained entity knowledge to distinguish visually similar entities while performing accurate grounding.

\section{KVG-Bench}
\label{sec:task}

\subsection{Task Definition}
\label{sec:task_def}

\begin{figure*}[t]
    \centering
    \includegraphics[width=\linewidth]{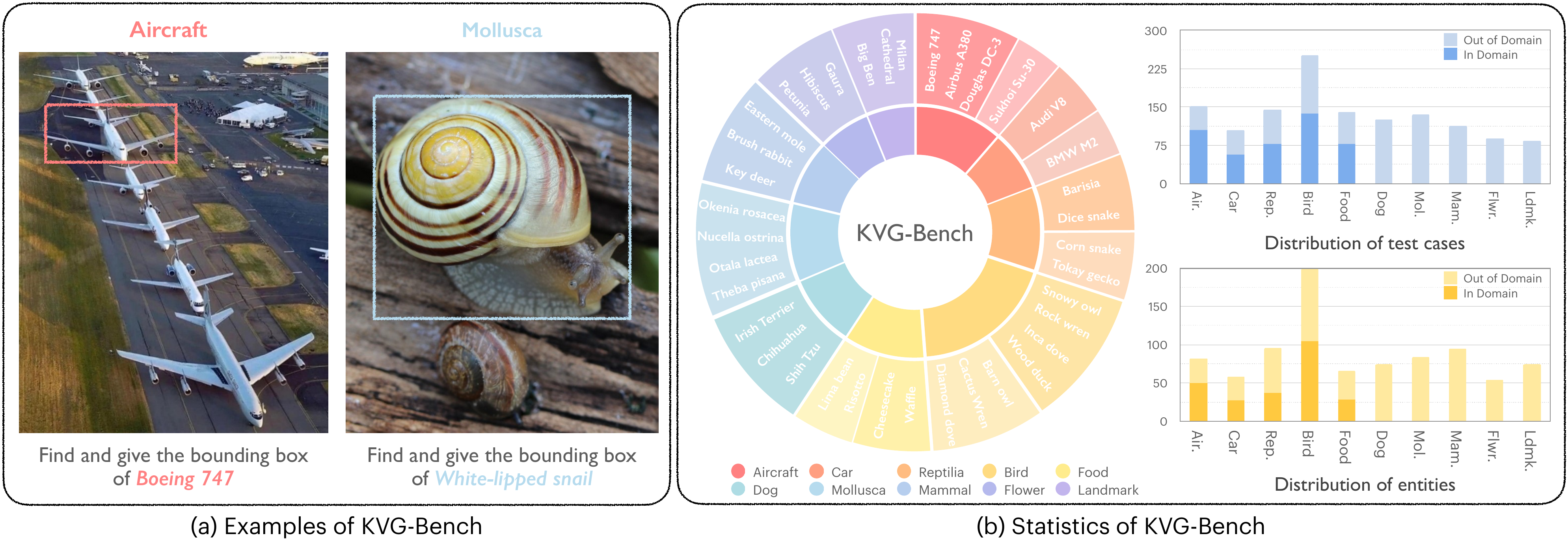}
    \caption{(a) \ourbench images contain multiple subordinate-category entities (e.g., multiple Boeing models in the left image); (b) \ourbench exhibits high diversity across categories and entities.}
    \label{fig:benchmark}
\end{figure*}

The task of knowledge-intensive visual grounding (KVG) aims to predict the bounding box $B = f_\theta(X_I, X_T)$ of a target entity through the joint understanding of visual input $X_I$ and textual query $X_T$. 
While sharing structural similarities with referring expression comprehension (REC), the KVG task significantly elevates the challenge beyond standard REC tasks.
As exemplified in \cref{fig:benchmark}a, the queries of KVG involve fine-grained entity specifications (e.g., ``Boeing 747'' and ``White-lipped snail'') rather than generic categories such as ``aircraft'' and ``mollusk''.
Moreover, each image typically contains multiple objects belonging to the same category as the target entity (e.g., several aircraft within a single image). 
This setup requires models to leverage fine-grained entity knowledge and perform careful visual comparison to identify the correct grounding target among visually similar entities.



\subsection{Benchmark Construction}
\label{sec:task_bench}

\ourbench comprises 1,336 test instances spanning 10 categories with 882 distinct entities, as statistically visualized in \cref{fig:benchmark}b.
The construction includes two key parts: image collection and data annotation.

We designed a meticulous collection process to ensure the diversity and complexity of the images.
First, we carefully selected 10 categories from the field of fine-grained visual recognition (FGVR)~\cite{welinder2010caltech, maji2013fine, krause20133d, stanforddogs, nilsback2008automated} that are suitable for visual grounding, excluding categories which are challenging for object localization such as ``sports'' and ``scene''.
Second, an entity list for each category was systematically developed through initial extraction of fine-grained labels from existing datasets, followed by comprehensive enrichment of entity names via ChatGPT-assisted expansion.
We then retrieved web images using these entity names as search queries, enforcing strict quality criteria: each image must contain at least two entities from the same category with clear visual disparities. 

The annotation process prioritized quality control.
Five annotators independently annotated each image with bounding boxes and entity labels by cross-referencing contextual information (e.g. caption, webpage metadata) with authoritative sources (e.g., Wikipedia entries) to verify entity identities.
To ensure consistency, all the annotations underwent independent re-evaluation by annotators who did not participate in the initial labeling, with conflicting cases cross-verified through multi-annotator reconciliation and persistently inconsistent instances eliminated to ensure annotation accuracy.
By integrating comprehensive validation protocols and expert-aligned annotation workflows, \ourbench provides a challenging testbed for evaluating knowledge-intensive visual grounding in MLLMs.

To evaluate generalization, we divide the benchmark into \textit{seen} and \textit{unseen} splits based on whether the target entity’s category appears in the training data. 
The seen split contains 5 categories (Aircraft, Car, Reptilia, Bird, Food), while the unseen split includes 5 held-out categories (Dog, Mollusca, Mammal, Flower, Landmark). 
This design evaluates both in-domain grounding and out-of-domain generalization to novel fine-grained concepts.

\subsection{Human Evaluation}
\label{sec:task_human}

To assess the difficulty of \ourbench, we conducted human evaluations with 11 non-expert volunteers under two experimental settings: Closed-Book (no external resources) and Open-Book (allowing participants to consult Wikipedia once per instance to simulate expert-level knowledge integration).
Participants were randomly assigned several categories, with each category being evaluated by at least five evaluators to mitigate knowledge bias.
The evaluation results, as shown in \cref{tab:main_result}, reveal significant performance differences between settings.
Notably, the Open-Book Setting demonstrated significant performance elevation (78.83\% accuracy) compared to Closed-Book results (56.41\%).
This validates that \ourbench requires synergistic integration of expert-level knowledge and fine-grained visual comparison, thereby highlighting the importance of integrating domain knowledge with visual grounding.

\section{Method}
\label{sec:method}

\begin{figure*}[t]
    \centering
    \includegraphics[width=\linewidth]{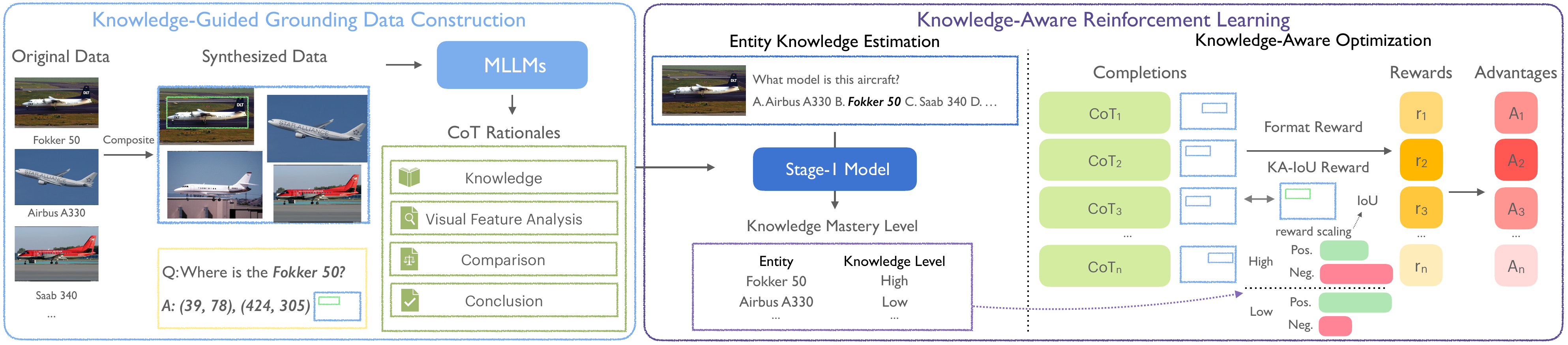}
    \caption{Overview of the proposed two-stage knowledge-aware training framework. (Left) Knowledge-guided grounding data construction and CoT-SFT training, which establish knowledge-guided reasoning for visual grounding. (Right) KARL, which estimates entity-level knowledge mastery and performs knowledge-aware reward scaling to balance positive and negative optimization signals under GRPO.}
    \label{fig:method}
\end{figure*}

While current MLLMs exhibit strong fine-grained entity knowledge and grounding capabilities, we observe a notable knowledge–grounding gap in the KVG task.
In such scenarios, models may correctly identify fine-grained entities, yet fail to reliably ground the same entities in the image.
This indicates that current training paradigms do not sufficiently align entity-level knowledge utilization with visual grounding optimization.
To address this limitation, we propose a two-stage knowledge-aware training framework (\cref{fig:method}) that explicitly encourages effective knowledge utilization during grounding while maintaining training stability.
\\

\subsection{Knowledge-Guided Grounding Data Construction}
\label{sec:method_data}
To bridge the knowledge–grounding gap, we construct training data that synthesize multi-entity grounding scenarios and augments them with knowledge-guided chain-of-thought (CoT) rationales.
This design alleviates the scarcity of suitable grounding data while providing a reasoning-aware initialization that encourages models to leverage entity knowledge during grounding. \\
\textbf{Multi-Entity Grounding Data Synthesis.}
Training images must satisfy two criteria:
(1) fine-grained entity annotations with precise bounding boxes, and
(2) multiple visually similar entities within the same image.
To address the scarcity of such data, we leverage multiple FGVR datasets across five categories as the source of entity annotations and corresponding images.
Since the original datasets lack bounding box annotations, we design a two-step annotation procedure.
We first use Qwen2-VL-7B~\cite{wang2024qwen2} to generate bounding boxes via entity-specific prompts, and then selectively refine predictions using SAM3~\cite{carion2025sam3segmentconcepts} to improve annotation quality.
Manual inspection on a random subset of 200 samples indicates over 95\% accuracy, where a prediction is considered correct if the generated box achieves an IoU greater than 0.5 with manually verified annotations.
To further increase grounding difficulty and prevent shortcut learning, we synthesize composite images containing at least two entities from the same category using structured layouts, while preserving annotation consistency.
This design encourages models to perform fine-grained discrimination rather than relying on coarse memorization.
Importantly, the synthesized training data is constructed from external FGVR datasets and is fully disjoint from \ourbench, ensuring zero image overlap during training.\\
\textbf{Knowledge-Guided Reasoning Generation.}
The primary objective of the first-stage training is to enable the model to utilize its internal entity knowledge for visual grounding, thereby narrowing the knowledge–grounding gap observed in KVG.
Rather than directly optimizing grounding accuracy, this stage encourages the model to incorporate entity knowledge into the reasoning process.
To construct such reasoning data, we employ Qwen2-VL-72B~\cite{wang2024qwen2} and Qwen3-VL-32B~\cite{bai2025qwen3} as teacher models to generate knowledge-guided CoT rationales.
The teachers are provided with the image, the target entity label, and the ground-truth bounding box, together with a prompt that instructs the model to produce step-by-step reasoning before predicting the final grounding result.
The prompt encourages the rationale to explicitly reference entity-related knowledge, including comparisons of fine-grained visual attributes, distinguishing characteristics among visually similar entities, and verification of these attributes against visual evidence in the image.
Example prompts are briefly illustrated in the main text, while full implementation details are provided in the supplementary material.

These rationales illustrate how entity knowledge can be invoked and verified against visual evidence before producing the final grounding prediction.
We then perform supervised fine-tuning using the constructed CoT rationales, which we refer to as CoT-SFT.
The resulting stage-1 model demonstrates a preliminary ability to incorporate entity knowledge during reasoning, serving as a stable initialization for subsequent knowledge-aware reinforcement learning.

\subsection{Knowledge-Aware Reinforcement Learning}
\label{sec:method_karl}
Following the first-stage training, we optimize the model through reinforcement learning to further enhance the model’s knowledge-intensive visual grounding capability, building upon its acquired knowledge-guided reasoning foundation.
However, models typically exhibit heterogeneous knowledge mastery across different entities. 
Applying uniform reward signals to all samples may disproportionately reinforce entities that are already well mastered while providing insufficient learning pressure for under-mastered ones. 
Such imbalance can further widen entity-level performance disparities and lead to suboptimal optimization.
Motivated by this observation, we propose Knowledge-Aware Reinforcement Learning (KARL), which modulates reward signals according to the model’s estimated entity-level knowledge mastery.\\
\textbf{Entity Knowledge Estimation.}
Since grounding performance may depend on multiple factors beyond entity knowledge, we estimate knowledge mastery using a recognition task that isolates the model's knowledge of fine-grained entities.
Let $\mathcal{E}$ denote the set of entities in the training data. 
For each entity $e \in \mathcal{E}$, we estimate the model’s knowledge mastery (i.e., its ability to correctly identify the entity) using single-entity images sampled from the source dataset. 
Specifically, we randomly sample a set of images $\{x_j^{(e)}\}_{j=1}^{N_e}$ containing entity $e$ and convert each image into a multiple-choice entity recognition question. 
Given an image $x_j^{(e)}$, the model is required to select the correct fine-grained entity label among several distractors drawn from the same category.
The knowledge mastery of entity $e$ is estimated using its average accuracy over sampled images. 
Based on this estimate, each entity is assigned to a discrete knowledge level via a mapping $k_e = \phi(e)$, where $k_e \in \{1, \ldots, K\}$ denotes the knowledge level of entity $e$.
Implementation details are provided in the supplementary material.\\
\textbf{Knowledge-Aware Optimization.}
To account for heterogeneous entity knowledge mastery, we modulate the reinforcement learning reward according to the knowledge level $k_e$ assigned to the entity $e$ in each sample. 
Our reward design consists of a knowledge-aware Intersection over Union (IoU) reward together with a format reward.
The KA-IoU reward evaluates the spatial alignment between the predicted bounding box $\tilde{B}$ and the ground-truth box $B$, while incorporating entity-level knowledge modulation. 
It is defined as
\begin{equation}
R_{\mathrm{KA\text{-}IoU}} =
\begin{cases}
s^{+}_{k_e} \cdot \mathrm{IoU}(B, \tilde{B}), & \text{if } \mathrm{IoU}(B, \tilde{B}) \ge \tau, \\
s^{-}_{k_e}, & \text{otherwise},
\end{cases}
\end{equation}
where $\tau$ is a threshold for valid spatial alignment. 
The positive scaling factor $s^{+}_{k_e}$ and negative scaling factor $s^{-}_{k_e}$ are determined by the entity’s knowledge level. 
Specifically, entities with higher knowledge levels receive smaller positive scaling and stronger negative scaling (i.e., more negative rewards for incorrect predictions), while entities with lower knowledge levels receive larger positive scaling and weaker negative scaling.
In addition to grounding accuracy, we include a lightweight format reward to enforce valid output structure. 
Specifically, regular expression-based pattern matching is used to ensure that reasoning traces are enclosed within ``<think>'' and ``</think>'' tags, and that the final grounded prediction is enclosed within ``<answer>'' and ``</answer>'' tags with a valid bounding box format. 
This auxiliary reward encourages structurally consistent outputs during optimization.
The overall reward is defined as:  $R = R_{\mathrm{KA\text{-}IoU}} + R_{\mathrm{format}}$.

We optimize the policy using Group Relative Policy Optimization (GRPO)~\cite{guo2025deepseek}. 
For each grounding query $q$, a group of candidate outputs is sampled from the old policy, and policy updates are performed using normalized group-relative advantages with Kullback–Leibler (KL) regularization. \\
\textbf{Data Filtering.}
To improve training efficiency, we filter data using the Stage-1 CoT-SFT model before Stage-2 reinforcement learning. 
For each training instance, we perform REC inference 4 times with stochastic sampling. 
Instances with 4/4 correct predictions are considered trivial and removed, while those with 0/4 correct predictions are regarded as overly difficult and also discarded. 
We retain only samples with mixed outcomes (1–3 correct predictions), resulting in approximately 2K instances used for Stage-2 reinforcement learning.
\section{Experiment}
\label{sec:exo}

\subsection{Implementation Details}
\label{sec:exp_detail}
\begin{table*}[t]
\caption{KVG results of \ours and baseline models. \ours achieves the best overall performance among all models, with particularly strong improvements on unseen categories.}
\label{tab:main_result}
\centering
\resizebox{\textwidth}{!}{
\begin{tabular}{l|ccccc|c|ccccc|c|c}
\toprule

\multirow{2}{*}{\textbf{Models}} & \multicolumn{6}{|c}{\textbf{Seen Categories}} & \multicolumn{6}{|c|}{\textbf{ Unseen categories }} & \multirow{2}{*}{\textbf{Avg.}} \\ 
\cmidrule(lr){2-7} \cmidrule(lr){8-13} 
 & \textbf{Air.} & \textbf{Car} & \textbf{Rep.} & \textbf{Bird} & \textbf{Food} & \textbf{Avg.} & \textbf{Dog} & \textbf{Mol.} & \textbf{Mam.} & \textbf{Flwr.} & \textbf{Ldmk.} & \textbf{Avg.} & \\
 
\midrule
\multicolumn{14}{c}{\textbf{Human Evaluation}}\\
\midrule 
\rowcolor{lightergray}
Human & 59.33 & 66.67 & 50.84 & 44.17 & 65.33 & 57.27 & 48.33 & 45.33 & 51.67 & 64.45 & 68.00 & 55.56 & 56.41 \\
\rowcolor{lightergray}
Human + search & 81.33 & 85.56 & 68.00 & 74.17 & 86.67 & 78.03 & 78.89 & 74.00 & 74.17 & 84.44 & 86.67 & 79.63 & 78.83 \\

\midrule 
\multicolumn{14}{c}{\textbf{Large-Scale MLLMs}}\\
\midrule
\rowcolor{lightergray}
InternVL2-76B~\cite{chen2024expanding} & 62.50 & 74.04 & 60.00 & 41.04 & 76.43 & 59.22 & 78.40 & 51.11 & 56.25 & 43.82 & 55.42 & 57.90 & 58.68 \\
\rowcolor{lightergray}
Qwen2-VL-72B~\cite{wang2024qwen2} & 63.16 & 75.96 & 59.31 & 40.24 & 77.14 & 59.34 &  80.80 & 42.96 & 59.82 & 65.17 & 66.27 & 62.32 & 60.55  \\
\rowcolor{lightergray}
Qwen3-VL-32B~\cite{bai2025qwen3} & 72.37 & 68.27 & 63.45 & 48.21 & 77.14 & 63.38 & 87.20 & 48.89 & 51.79 & 66.29 & 81.93 & 66.18 & 64.52 \\

\midrule 
\multicolumn{14}{c}{\textbf{Specialist Grounding Models}}\\
\midrule

YOLO-World~\cite{Cheng2024YOLOWorld} & 41.45 & 28.85 & 8.28 & 14.74 & 30.71 & 23.36 & 50.40 & 2.22 & 24.11 & 1.12 & 3.61 & 17.83 & 21.11 \\

G-DINO-1.6-Pro~\cite{groundingdino_1_6} & 39.47 & 41.35 & 48.97 & 23.11 & 24.29 & 33.59 & 44.00 & 40.00 & 39.29 & 32.58 & 27.71 & 37.68 & 35.25 \\

DINO-X~\cite{ren2024dinoxunifiedvisionmodel} & 43.42 & 49.04 & 42.76 & 28.29 & 41.43 & 38.89 & 62.40 & 35.56 & 48.21 & 31.46 & 49.40 & 45.77 & 41.69 \\

\midrule 
\multicolumn{14}{c}{\textbf{7B-Scale MLLMs}}\\
\midrule

Shikra-7B~\cite{chen2023shikra} & 20.39 & 25.96 & 15.17 & 16.33 & 28.57 & 20.33 & 51.20 & 19.26 & 25.00 & 16.85 & 22.89 & 27.94 & 23.43 \\
CogVLM-G~\cite{wang2024cogvlm} & 46.71 & 64.42 & 49.66 & 34.26 & 63.57 & 48.61 & 79.20 & 31.11 & 54.46 & 56.18 & 66.27 & 56.43 & 51.80 \\
DeepSeek-VL2~\cite{wu2024deepseek} & 51.32 & 60.57 & 53.10 & 29.08 & 63.57 & 47.98 & 62.40 & 35.56 & 50.89 & 44.94 & 39.76 & 47.06 & 47.60 \\
Qwen2-VL-7B~\cite{wang2024qwen2} & 48.03 & 74.04 & 51.30 & 33.07 & 65.71 & 50.38 & 76.00 & 33.33 & 54.46 & 57.30 & 59.04 & 55.33 & 52.40 \\
Qwen3-VL-8B~\cite{bai2025qwen3} & 61.84 & 68.27 & 55.17 & 40.24 & 75.00 & 56.94 & 73.60 & 44.44 & 42.86 & 65.17 & 66.27 & 57.54 & 57.19 \\
\midrule 
\multicolumn{14}{c}{\textbf{Reasoning MLLMs}}\\
\midrule 
Perception-R1~\cite{yu2025perceptionr} & 23.03 & 32.69 & 14.48 & 12.35 & 37.14 & 21.84 & 41.60 & 11.85 & 26.79 & 16.85 & 28.92 & 25.18 & 23.20 \\
Visual-RFT~\cite{liu2025visual} & 36.18 & 51.92 & 32.41 & 20.70 & 47.86 & 34.72 & 52.80 & 21.48 & 31.25 & 46.07 & 44.58 & 38.24 & 36.15 \\
UniVG-R1~\cite{bai2025univg} & 57.24 & 74.04 & 52.41 & 35.06 & 78.57 & 55.30 & 73.60 & 40.00 & 55.36 & 60.67 & 67.46 & 58.46 & 56.59 \\

\midrule 
\rowcolor{lighterpurple}
\ours & \textbf{71.05} & \textbf{83.65} & \textbf{70.34} & \textbf{45.02} & \textbf{87.86} & \textbf{67.30} & \textbf{85.60} & \textbf{50.37} & \textbf{69.64} & \textbf{80.90} & \textbf{67.47} & \textbf{70.04} & \textbf{68.41} \\

\bottomrule
\end{tabular}
}
\end{table*}
\textbf{Datasets.}
We conducted experiments based on several FGVR datasets including FGVC-Aircraft~\cite{maji2013fine}, Stanford-Cars~\cite{krause20133d}, iNaturalist2017~\cite{van2018inaturalist},  and Food101~\cite{bossard2014food}.
For the iNaturalist2017 dataset, we use the categories of \textit{Reptilia} and \textit{Aves} as the training sources for constructing grounding data.
Following the pipeline described in Sec.~\ref{sec:method_data}, we construct 25K training samples for Stage-1 and further select 2K instances for Stage-2 reinforcement learning. \\
\textbf{Baseline Models.}
We build \ours upon Qwen3-VL-8B-Instruct~\cite{bai2025qwen3} (hereafter Qwen3-VL-8B) due to its strong visual grounding capability and rich knowledge. 
For fair comparison, we evaluate several MLLMs with strong grounding ability across different model scales.
For large-scale models, we used InternVL2-Llama3-76B~\cite{chen2024expanding}, Qwen2VL-72B~\cite{wang2024qwen2}, and Qwen3VL-32B-Instruct~\cite{bai2025qwen3} (hereafter Qwen3-VL-32B). 
For 7B-scale models, we used Shikra~\cite{chen2023shikra}, CogVLM-Grounding~\cite{wang2024cogvlm}, DeepSeek-VL-2~\cite{wu2024deepseek}, Qwen2-VL-7B~\cite{wang2024qwen2} and Qwen3-VL-8B~\cite{bai2025qwen3}.
Additionally, we conducted comparisons with three specialist models: YOLO-World~\cite{Cheng2024YOLOWorld}, Ground-ingDINO-1.6-Pro~\cite{groundingdino_1_6} and DINO-X~\cite{ren2024dinoxunifiedvisionmodel}. Moreover, We also compare with recent reasoning-guided grounding MLLMs: Perception-R1~\cite{yu2025perceptionr}, Visual-RFT~\cite{liu2025visual} and UniVG-R1~\cite{bai2025univg}. For Qwen3-VL-8B, we further evaluate several variants under different reasoning and training settings. \\
\textbf{Evaluation Settings.} 
We evaluate all methods on \ourbench using accuracy as the evaluation metric, following standard practice in referring expression comprehension (REC)~\cite{you2024ferret, chen2023shikra, wang2024qwen2}.
Specifically, given a predicted bounding box $\tilde{B}$ and ground-truth bounding box $B$, we compute their IoU, and a prediction is considered correct if $\text{IoU} \geq 0.5$.
For multi-category evaluation, the overall accuracy is computed as a weighted average according to the number of instances in each category. \\
\textbf{Training Details.}
All experiments are conducted on 8 NVIDIA A800 80GB GPUs. 
In Stage 1, we optimize the model using Adam~\cite{adam} with a learning rate of $1\times10^{-6}$ and $\beta_1=0.9$, $\beta_2=0.999$. 
In Stage 2, we adopt AdamW~\cite{loshchilov2018decoupled} with a learning rate of $5\times10^{-7}$ and the same momentum parameters. 
The accumulated batch size is set to 16 in stage 1 and 8 in stage 2.
For GRPO training, the maximum completion length is set to 1500 tokens, and four samples are generated for each input query. 
In KARL, entities are grouped into five knowledge levels according to the estimated knowledge mastery. 
The positive reward scaling factors are set to 0.65, 0.85, 1.0, 1.2, and 1.4 (capped at 1.0), while the negative reward scaling factors are set to $-0.6$, $-0.5$, $-0.4$, $-0.2$, and $0.0$.

\begin{table}[t]
\caption{Performance Comparison of \ours and Qwen3VL-8B on general multimodal benchmarks. KARL maintains performance comparable to the base model, indicating that general multimodal capabilities are preserved.}
\label{tab:general}
\centering
\resizebox{0.7\linewidth}{!}{
\begin{tabular}{l|cccc}
\toprule

\bf Models & \textbf{MMStar} & \textbf{RealWorldQA} & \bf AI2D & \bf CV-Bench \\

\midrule 

Qwen3VL-8B\tablefootnote{reproduced using lmms-eval~\cite{zhang2024lmmsevalrealitycheckevaluation,lmms_eval2024}} & 63.5 & 69.4 & 83.6 & 86.1 \\
\ours & 64.1 & 69.9 & 83.7 & 86.7 \\
\bottomrule
\end{tabular}
}
\end{table}
\subsection{Main Results}
\label{sec:exp_main}
\Cref{tab:main_result} presents the performance of \ours and baseline models on \ourbench. 
Overall, \ours consistently outperforms all baselines across both seen and unseen categories, demonstrating the effectiveness of knowledge-aware reinforcement learning for knowledge-intensive visual grounding.

On \textbf{seen categories}, \ours reaches \textbf{67.30\%}, outperforming strong 7B baselines such as Qwen3-VL-8B (56.94\%) and even surpassing several large-scale MLLMs (e.g., 63.38\% for Qwen3-VL-32B).
This result suggests that explicitly encouraging models to utilize fine-grained entity knowledge during grounding can yield improvements beyond those obtained from model scaling alone.
More importantly, on \textbf{unseen categories}, which evaluate cross-domain generalization, \ours achieves the best performance of \textbf{70.04\%}.  
It consistently surpasses both large-scale MLLMs and recent reasoning-guided grounding models such as UniVG-R1 (58.46\%) and Visual-RFT (38.24\%).
The improvements are observed across diverse semantic domains, including Dog (85.60\%), Flower (80.90\%), and Mammal (69.64\%), indicating that the model can effectively transfer entity-level knowledge to previously unseen categories.
\begin{table}[t]
\caption{Ablation study on the effectiveness of the proposed two-stage knowledge-aware training framework. The results show that the two-stage design and the KARL optimization consistently improve both seen and unseen performance.}
\label{tab:ablation-stage2}
\centering
\begin{tabular}{cl|cc|c}
\toprule

\bf Stage & \bf Method & \bf Seen \  & \bf Unseen & \bf Overall \\

\midrule 

\multirow{2}{*}{1} & SFT & 61.49 & 62.87 & 62.05 \\
& CoT-SFT & 67.05 & 66.73 & 66.92 \\
\midrule
\multirow{4}{*}{2} & SFT & 59.22 & 61.76 & 60.25 \\
& CoT-SFT & 67.05 & 65.07 & 66.24 \\
& GRPO & 67.29 & 66.36 & 66.92 \\
\rowcolor{lighterpurple}
& KARL & \bf 67.30 & \bf 70.04 & \bf 68.41 \\
\bottomrule
\end{tabular}
\end{table}
Notably, generic reasoning strategies do not consistently improve performance on KVG. 
For example, adding zero-shot CoT prompting to the base model yields performance comparable to the original model (see~\cref{fig:teaser}b). 
Similarly, recent reasoning-guided grounding methods such as Visual-RFT and UniVG-R1, both built upon Qwen2-VL-7B, show only limited improvements over their base model and do not consistently outperform it on \ourbench, with performance in some cases even lower. 
These observations suggest that generic reasoning strategies are insufficient for knowledge-intensive grounding.
In contrast, by explicitly encouraging models to activate entity-level knowledge during grounding and adapting reinforcement learning signals according to knowledge mastery, \ours achieves the best overall performance of \textbf{68.41\%}, establishing a new state of the art on \ourbench. \\
\textbf{General Capability.}
To comprehensively assess the model’s general capabilities across diverse multimodal scenarios, we conducted evaluations on established multimodal benchmarks.
As shown in \cref{tab:general}, \ours maintains performance comparable to the base model, demonstrating preserved general capabilities without degradation.

\subsection{Ablation Study}
\label{sec:exp_ablation}
\begin{table}[t]
\caption{Ablation study on different reward scaling strategies. Removing either positive or negative reward scaling degrades performance, highlighting the importance of knowledge-aware reward signals.}
\label{tab:ablation-rewards}
\centering
\begin{tabular}{l|cc|c}
\toprule

\textbf{Strategy} \ & \textbf{Seen} \  & \textbf{Unseen} & \textbf{Overall} \\

\midrule
\rowcolor{lighterpurple}
KARL & \bf 67.30 & \bf 70.04 & \bf 68.41 \\
w/o Neg. & 66.67 & 66.91 & 66.77 \\
w/o Pos. & 67.17 & 66.36 & 66.84 \\
w/o Pos. \& Neg. & 67.29 & 66.36  & 66.92 \\

\bottomrule
\end{tabular}
\end{table}

To evaluate the effectiveness of the proposed two-stage knowledge-aware training framework (KARL) and the contributions of its individual components, we conducted a series of ablation experiments. The analysis focuses on the necessity of the two-stage framework and the impact of knowledge-aware reward shaping strategies. \\
\textbf{Effectiveness of Training Strategies Across Stages.} 
As shown in \cref{tab:ablation-stage2}, we compare different training strategies across the two stages.
Replacing standard SFT with CoT-SFT in Stage 1 improves performance on seen categories from 61.49\% to 67.05\%, demonstrating that knowledge-guided reasoning supervision provides a stronger initialization for knowledge-intensive grounding.

In Stage 2, we further optimize the Stage-1 models using different training strategies on the same filtered training data. Specifically, Stage-2 SFT continues training from the Stage-1 SFT model, while all other methods (CoT-SFT, GRPO, and KARL) are initialized from the same Stage-1 CoT-SFT checkpoint.
Comparing Stage-2 results, both GRPO and KARL further improve performance over CoT-SFT across seen and unseen categories, highlighting the benefit of reinforcement learning for refining grounding predictions.
More importantly, compared with standard GRPO, KARL achieves comparable performance on seen categories (67.30\% vs.\ 67.29\%) while delivering a substantial improvement on unseen categories (70.04\% vs.\ 66.36\%). 
This result suggests that knowledge-aware reward scaling helps rebalance optimization across entities with different knowledge levels, leading to improved generalization on previously unseen categories.\\  
\textbf{Analysis of Reward Scaling Strategies.} 
To further analyze the contributions of KARL, we examine the effects of different reward scaling components in \cref{tab:ablation-rewards}.
Specifically, "w/o Neg." applies only positive reward scaling; "w/o Pos." applies only negative penalty scaling; "w/o Pos. \& Neg. " is vanilla GRPO with fixed positive and negative rewards.
Removing the negative scaling (w/o Neg.) reveals the critical role of penalizing incorrect answers: performance drops on both seen (67.30\%→66.67\%) and unseen (70.04\%→66.91\%) categories, with a more pronounced decline on unseen ones, indicating that knowledge-guided penalties help the model avoid spurious correlations and enhance generalization. 
In contrast, removing positive scaling (w/o Pos.) highlights the importance of reinforcing correct predictions. 
When positive scaling is removed, unseen accuracy decreases substantially (70.04\%→66.36\%), indicating that reinforcing correct predictions is important for improving generalization to unseen categories.

Finally, when comparing KARL with the vanilla GRPO baseline (w/o Pos. \& Neg.), unseen performance suffers a large degradation (70.04\%→66.36\%), confirming that the combination of positive and negative knowledge scaling is essential for achieving strong generalization beyond memorization. \\
\textbf{Overall}, these results validate the effectiveness of the proposed knowledge-aware training framework and reward scaling strategy in KARL, which improves grounding performance and generalization to unseen categories.

\subsection{Analysis}
\label{sec:exp_analysis}
\begin{figure}[t]
    \centering
    \includegraphics[width=.8\linewidth]{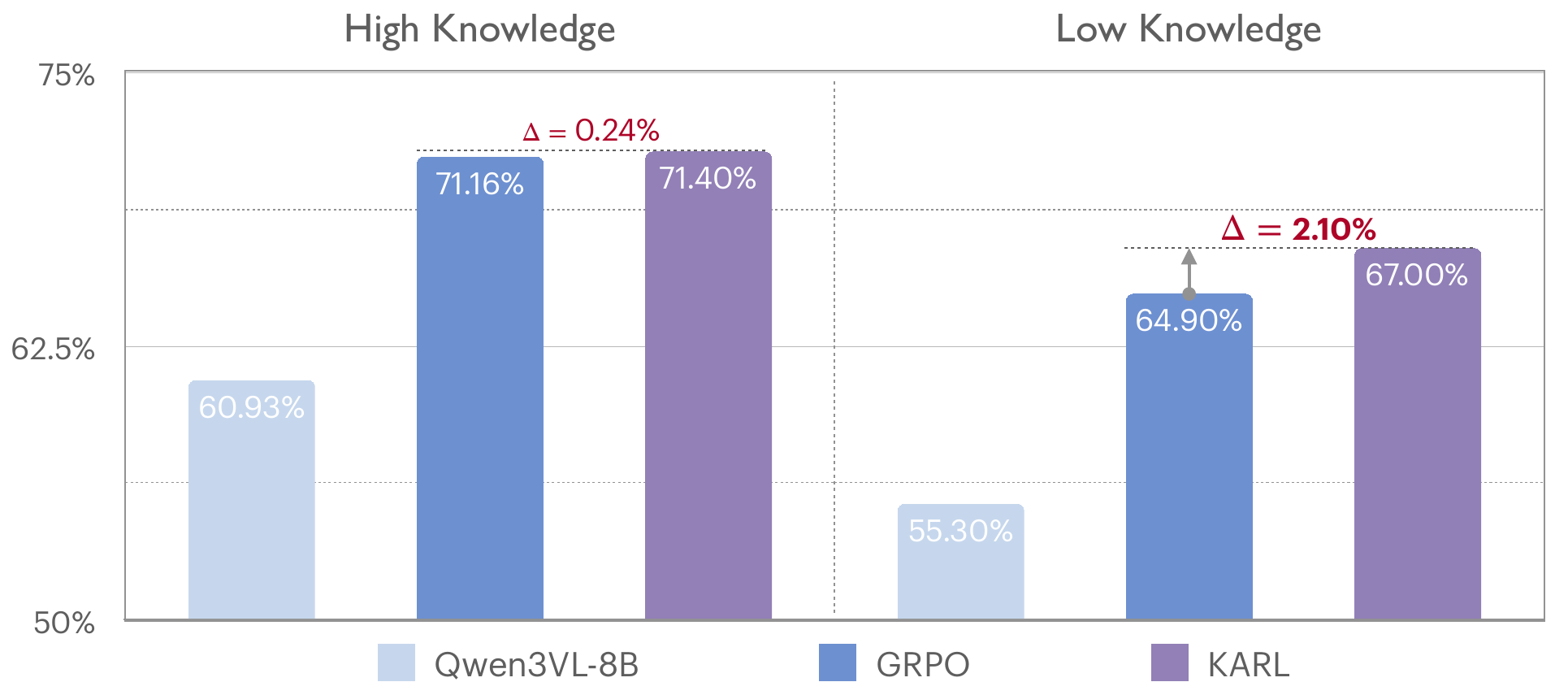}
    \caption{Knowledge evaluation results. \ours shows larger gains on low-knowledge entities, demonstrating the effectiveness of knowledge-aware reward scaling on low knowledge.}
    \label{fig:knowledge}
\end{figure}
\begin{figure}[t]
    \centering
    \includegraphics[width=\linewidth]{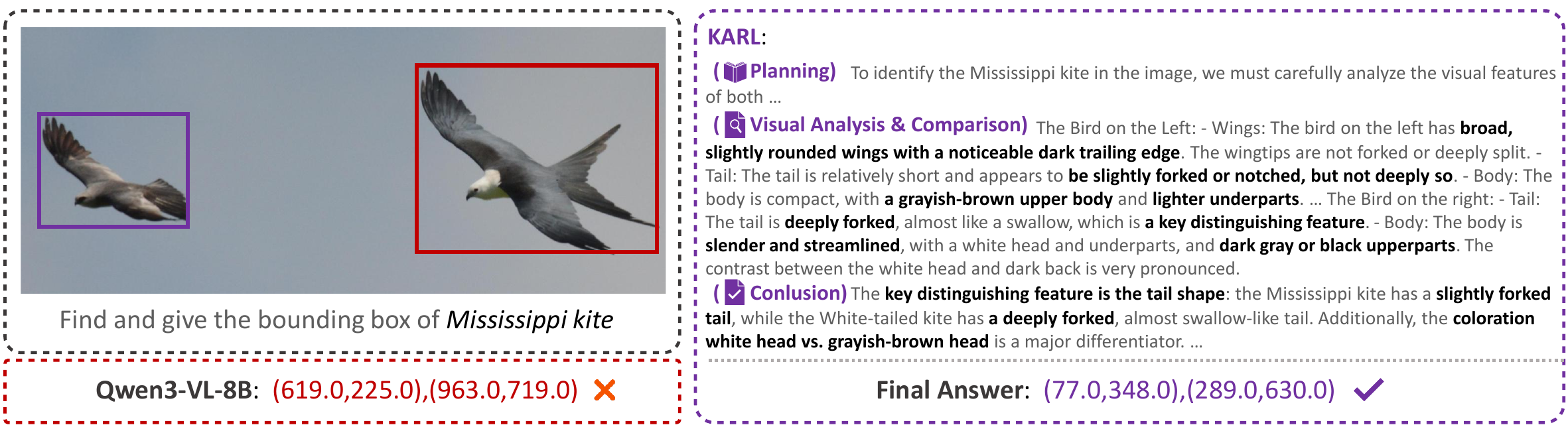}
    \caption{Case study comparing \ours and Qwen3-VL-8B. KARL correctly grounds the target entity by leveraging knowledge-guided reasoning, while the baseline model produces an incorrect prediction.}
    \label{fig:case}
\end{figure}

To gain deeper insights into our model's capabilities, we conducted the following analysis, examining how knowledge-aware reward scaling affects performance across different knowledge conditions and how it influences the model’s reasoning behavior in practice: \\
\textbf{Knowledge-Level Analysis.}
To simplify the analysis, we divide entities into two knowledge levels based on the estimated knowledge mastery: the top three categories are treated as high knowledge, while the remaining two are treated as low knowledge.
As shown in \cref{fig:knowledge}, both GRPO and \ours bring substantial improvements over the base model across different knowledge levels, indicating that reinforcement learning generally improves grounding performance regardless of the knowledge level. 
However, on low-knowledge entities, \ours achieves a much larger gain (+2.10\%) than GRPO, which is nearly nine times larger than the improvement observed in the high-knowledge setting. This result suggests that \textbf{knowledge-aware reward scaling in \ours enables more effective learning under low knowledge conditions}.\\

\textbf{Qualitative Results.} 
As illustrated in~\cref{fig:case}, we conducted a comparative case analysis between KARL and Qwen3-VL-8B. In this case, KARL follows a structured reasoning pipeline—initiating with a planning step, then performing a detailed visual analysis and comparison of the entities, and finally drawing a conclusion based on the most discriminative features (e.g., slightly forked vs. deeply forked), thereby arriving at the accurate identification. The visual evidence demonstrates our model’s capability to generate accurate answers through a knowledge-aware reasoning process that systematically integrates domain specific knowledge with visual observations, in contrast to the baseline model’s tendency to produce incorrect responses from superficial pattern recognition. \\

\section{Conclusion}
\label{sec:conclusion}

In this work, we investigate knowledge-intensive visual grounding and identify a gap between entity knowledge and grounding performance in current MLLMs. 
To address this limitation, we propose a knowledge-aware training paradigm that combines knowledge-guided reasoning supervision with knowledge-aware reinforcement learning. 
The resulting framework enables models to better incorporate entity-level knowledge into grounding decisions. 
Extensive experiments on \ourbench demonstrate that \ours consistently outperforms strong baseline models and achieves the best overall performance on the benchmark, while maintaining competitive performance on general multimodal tasks. 
Further analyses show that knowledge-guided reasoning provides a strong initialization for KVG, and that knowledge-aware reward scaling improves cross-domain generalization across entities with different knowledge levels. 
These results highlight the importance of explicitly integrating entity knowledge into the grounding process and suggest a promising direction for improving knowledge-intensive perception in MLLMs.

\bibliographystyle{splncs04}
\bibliography{main}

\newpage
\appendix
\title{Supplementary Materials}

\section{\ourbench}
\label{sec:appendix-bench}

This section details the data collection and annotation protocol for \ourbench.

\subsection{Data Collection}

\paragraph{Category Selection} 

Ten categories were strategically curated from established fine-grained visual recognition (FVGR) datasets including FGVC-Aircraft~\cite{maji2013fine}, Stanford-Cars~\cite{krause20133d}, iNaturalist2017~\cite{van2018inaturalist}, Food101~\cite{bossard2014food}, Stanford-Dogs~\cite{stanforddogs}, Flower-102~\cite{nilsback2008automated}, and Google-Landmarks-v2~\cite{weyand2020GLDv2}.
Categories requiring ambiguous spatial localization (e.g., "sports" and "scenes") were systematically excluded.

\paragraph{Entity List Curation}

We built the list of detailed entity names through a step-by-step process. 
First, we started with existing names from aforementioned FGVR datasets to make sure they fit the right categories.
Then, we used ChatGPT~\cite{gpt4} to collect more entities in these categories by querying with category name and example entities.
Finally, we checked all these entity names against Wikipedia to confirm their accuracy and avoid confusing or incorrect terms.

\paragraph{Web Image Retrieval}

The image collection process employed diversified search strategies, systematically generating query variations (e.g., ``X versus Y'', ``X compared to Y'', ``differences between X and Y'') to retrieve visually discriminative instances across search engines.
The image collection process focused on two core principles: diversity and challenge. 
To ensure diversity, images were collected to encompass a variety of distinct entities within each category, achieved through systematically varied search query combinations.
This approach guaranteed a wide range of entity interactions and visual scenarios.
For challenge, we specifically selected images containing multiple entities from the same category (requires fine-grained visual discrimination) or exhibited high visual similarity with subtle distinguishing features. 

\subsection{Annotation}

The annotation process prioritized quality control.
Five annotators manually annotated each image with bounding boxes and entity labels by cross-referencing contextual information (e.g. caption, webpage metadata) with authoritative sources (e.g., Wikipedia entries) to verify entity identities.
To ensure consistency, the annotations underwent independent re-evaluation by annotators who did not participate in the initial labeling, with conflicting cases cross-verified through multi-annotator reconciliation and persistently inconsistent instances eliminated to ensure annotation accuracy.

\section{Method}
\label{sec:appendix-method}

This section elaborates on the implementation details of the proposed method.

\subsection{Knowledge-Guided Grounding Data Construction}
\label{sec:appendix-method_data}

\paragraph{Multi-Entity Grounding Data Synthesis}

Our data processing pipeline comprises three core stages: categorization, fine-grained annotation, and image composition. \\
\textbf{Category Organization.}
We first organize images from external fine-grained visual recognition datasets into category-specific subsets according to their semantic domains.
Specifically, we use FGVC-Aircraft~\cite{maji2013fine}, Stanford-Cars~\cite{krause20133d}, and Food101~\cite{bossard2014food} as the data sources for the \textit{aircraft}, \textit{car}, and \textit{food} categories, respectively.
For iNaturalist2017~\cite{van2018inaturalist}, which covers multiple biological groups, we further select the \textit{Reptilia} and \textit{Aves} subsets to construct data for the \textit{reptilia} and \textit{bird} categories.
These five categories serve as the training sources for the grounding data used in our two-stage optimization pipeline. \\
\textbf{Box Annotation.}
Since the original datasets do not provide bounding box annotations, we adopt a two-step annotation procedure.
We first use Qwen2-VL-7B~\cite{wang2024qwen2} to generate bounding boxes with entity-specific prompts.
Given an image and its verified entity label $E$, we query the model with ``Find and give the bounding box of \{E\}'', and use the returned coordinates as pseudo annotations.
We then manually inspect the generated boxes on a random subset and identify categories with relatively lower annotation quality.
Based on this inspection, we further apply SAM3~\cite{carion2025sam3segmentconcepts} to selectively refine Qwen-generated predictions for those categories, rather than uniformly processing all data.
In practice, this refinement is mainly used for the \textit{food} category, where the initial predictions are comparatively less accurate.
Manual inspection on a random subset of 200 samples indicates over 95\% annotation accuracy, where a prediction is considered correct if its generated box achieves an IoU greater than 0.5 with manually verified annotations. \\
\textbf{Multi-Entity Image Composition.}
Based on each category-specific dataset $\mathcal{D}_c$, we synthesize multi-entity grounding examples by iteratively sampling instances without replacement.
At each iteration, we randomly sample $k$ images, where $k$ is drawn from a predefined discrete distribution over $[2,6]$.
To avoid trivial shortcuts, all sampled instances are required to correspond to different fine-grained entities.
The selected images are then composed into a single scene using one of four layout strategies: horizontal concatenation, vertical concatenation, grid arrangement, or random placement.
During composition, we apply the corresponding geometric transformations to remap each original bounding box to its new location in the synthesized image, thereby preserving annotation consistency.
The resulting composite dataset $\mathcal{D}_c^\prime$ contains multi-entity images, transformed bounding boxes, and entity labels, creating training examples in which correct grounding requires distinguishing among semantically related fine-grained entities rather than relying on coarse category cues.
Finally, we construct 25K samples for Stage-1 training and further select 2K instances for Stage-2 reinforcement learning.

\begin{figure*}
\centering
\begin{tcolorbox}[width=\textwidth,colback=black!5!white,colframe=black!75!black,title=CoT Generation Prompt]

<|vision\_start|>Image.jpg<|vision\_end|>

This image shows [entity\textsubscript{1}] ([bbox\textsubscript{1}]), [entity\textsubscript{2}] ([bbox\textsubscript{2}]), $\cdots$, and [entity\textsubscript{n}] ([bbox\textsubscript{n}]).

The bounding box of [target entity] is [target bbox].

Give the reasoning process that would identify it based on the image and your knowledge

Note that you MUST pay attention to the differences from other objects of the same type in this image and make a detailed comparison between them to find evidence that distinguishes this object from the others

Note that you MUST first analyze the visual features that help you make a judgment, and then compare the objects

Note that when an object is ``[Unknown]'', you can still make a comparison based on its visual features without knowing its name

\end{tcolorbox}
\end{figure*}

\paragraph{Knowledge-Guided Reasoning Generation}

\begin{figure*}[t]
    \centering
    \includegraphics[width=0.8\linewidth]{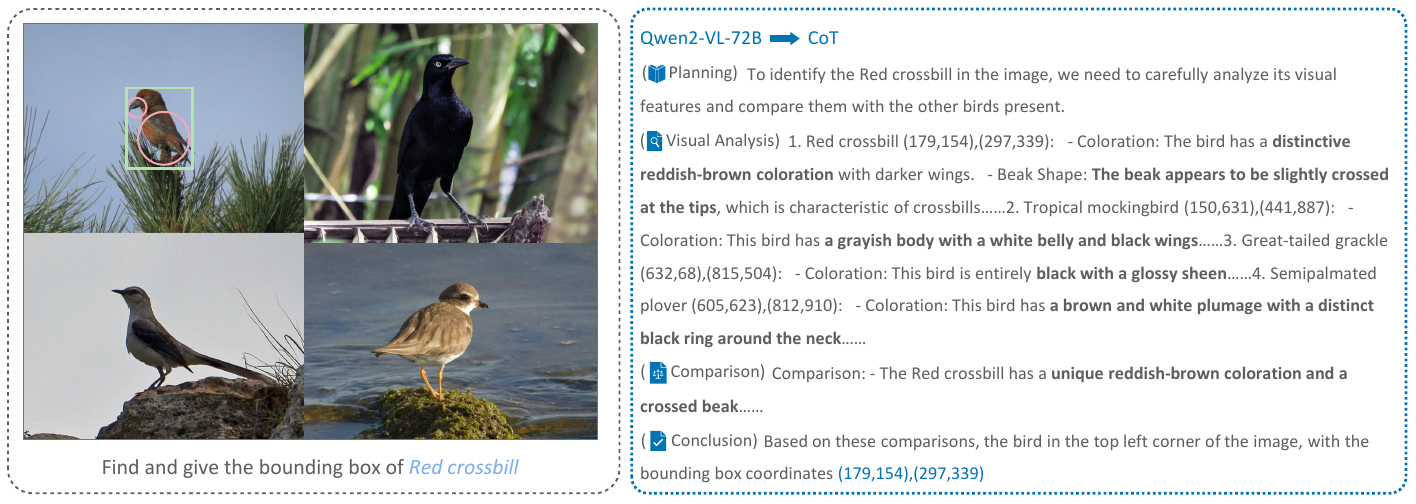}
    \caption{Example of Chain-of-Thought data generated by Qwen2-VL-72B. }
    \label{fig:case_cot}
\end{figure*}

To construct high-quality Chain-of-Thought (CoT) training data, we generate reasoning annotations by harnessing high-capacity open-source Multimodal Large Language Models (MLLMs).
Specifically, the model takes the image, ground-truth entity annotations (names and bounding boxes), and a CoT generation instruction as input, producing detailed reasoning processes for training. \cref{fig:case_cot} illustrates a representative training instance. 

To determine which model produces more effective CoT supervision, we further analyze the downstream SFT performance using CoT data generated by different models. As shown in \cref{tab:cot-sft-compare}, CoT data produced by \textbf{Qwen2-VL-72B-Instruct}~\cite{wang2024qwen2} leads to stronger SFT performance on the \textit{Aircraft}, \textit{Car}, and \textit{Reptilia} categories, while \textbf{Qwen3-VL-32B-Instruct}~\cite{bai2025qwen3} performs better on \textit{Bird} and \textit{Food}. This observation suggests that different models possess complementary strengths in capturing category-specific visual knowledge during reasoning generation. Based on this finding, we adopt a category-aware CoT data construction strategy. Specifically, we use Qwen2-VL-72B-Instruct to generate CoT annotations for Aircraft, Car, and Reptilia, and Qwen3-VL-32B-Instruct for Bird and Food, where each model demonstrates superior performance. This strategy enables us to leverage the complementary reasoning capabilities of different large-scale models, thereby improving the overall quality and diversity of the generated CoT training data.

\begin{table}[t]
\caption{Comparison of SFT performance using CoT data generated by different large-scale vision-language models.}
\centering
\resizebox{0.7\linewidth}{!}{
\begin{tabular}{l|ccccc}
\toprule
\textbf{Models} & \textbf{Air.} & \textbf{Car} & \textbf{Rep.} & \textbf{Bird} & \textbf{Food} \\
\midrule 
CoT-SFT (Qwen2VL-72B) & \bf 75.00 & \bf 76.92 & \bf 69.66 & 41.04 & 72.86  \\
CoT-SFT (Qwen3VL-32B) & 72.37 & 75.96 & 61.38 & \bf 44.62 & \bf 78.57  \\
\bottomrule
\end{tabular}
}
\label{tab:cot-sft-compare}
\end{table}

\subsection{Knowledge-Aware Reinforcement Learning}
\label{sec:method_karl}

\paragraph{Entity Knowledge Estimation}

To quantify the degree of visual knowledge possessed by the model, we construct a controlled visual recognition probe to estimate a knowledge mastery score for each entity. For every entity, we randomly sample up to 50 images containing the target entity and ask Qwen3-VL-8B-Instruct~\cite{bai2025qwen3} to identify its category. Each question is formulated as a four-choice multiple-choice task, where the candidate options are semantically similar categories drawn from the same super-category. The model’s average accuracy ($ACC_{avg}$) across the evaluated images is used as the knowledge mastery score of that entity.

We categorize entities into five knowledge levels based on $ACC_{avg}$: completely correct ($ACC_{avg} \geq 0.7$), mostly correct ($0.45 \leq ACC_{avg} < 0.7$), partially correct ($0.3 \leq ACC_{avg} < 0.45$), mostly incorrect ($0.1 \leq ACC_{avg} < 0.3$), and completely incorrect ($ACC_{avg} < 0.1$). These thresholds are designed with respect to the four-choice classification setting, where random guessing yields an expected accuracy of 25\%. In particular, the partially correct interval (0.30–0.45) corresponds to performance slightly above the chance level, while the mostly incorrect and completely incorrect ranges capture entities for which the model performs at or below chance level. Conversely, the mostly correct and completely correct ranges represent entities where the model demonstrates increasingly reliable recognition ability.

To obtain an unbiased estimation of visual knowledge distribution, we evaluate all entities in the dataset. The resulting statistics reveals a highly skewed knowledge landscape, Specifically, 46.79\% of entities are categorized as mostly incorrect and 19.27\% as completely incorrect, meaning that 66.06\% of entities cannot be reliably recognized by the model. An additional 21.36\% fall into the partially correct category, indicating unstable knowledge representations. In contrast, only a small fraction of entities are classified as mostly correct (11.40\%) or completely correct (1.18\%), suggesting that strong visual knowledge is concentrated on a very limited subset of concepts. These results indicate that visual knowledge in current MLLMs is highly uneven and sparse, motivating the need for knowledge-aware training strategies to better align entity-level knowledge with visual grounding.

\paragraph{Knowledge-Aware Optimization}
\label{sec:appendix-method_karl}

Table~\ref{tab:reward_scaling} presents the parameter search for the reward-scaling strategy used in our knowledge-aware reinforcement learning stage. Specifically, Scaling V1 adopts positive weights: [0.6,0.8,1.0,1.2,1.4] with relatively stronger negative penalties: [$-0.8$,$-0.6$,$-0.4$,$-0.2$,0.0], while Scaling V2 uses slightly different positive weights: [0.65,0.85,1.0,1.15,1.25] combined with milder penalties: [$-0.6$,$-0.5$,$-0.2$,$-0.1$,0.0]. The final KARL scaling integrates these two schedules into a unified configuration, resulting in positive weights: [0.65,0.85,1.0,1.2,1.4] and negative penalties: [$-0.6$,$-0.5$,$-0.4$,$-0.2$,0.0].

Combining the low-knowledge design from V1 with the high-knowledge design from V2, the final KARL scaling achieves the highest average performance (68.41\%) among all variants, demonstrating that integrating the two regimes effectively balances learning signals across entity types. This unified schedule is therefore selected as our reward-scaling strategy.

\section{Experiment}
\label{sec:appendix-exp}
\begin{table}[t]
\caption{Reward scaling parameter search for KARL. The proposed KARL scaling strategy achieves the best overall average performance}
\centering
\begin{tabular}{l|cc|c}
\toprule
\bf Strategy & \bf Avg \\
\midrule 
Scaling (KARL) \  & 68.41 \  \\
Scaling V1 \  & 67.73 \  \\
Scaling V2 \  & 67.29 \  \\

\bottomrule
\end{tabular}
\label{tab:reward_scaling}
\end{table}

This section provides additional details about the experiments and analysis, along with discussions on empirical findings.

\subsection{Implementation Details}
\label{sec:appendix-exp_detail}
\begin{table}[t]
\caption{Performance comparison on RefCOCO, RefCOCO+, and RefCOCOg benchmarks.}
\centering
\resizebox{\textwidth}{!}{
\begin{tabular}{l|ccc|ccc|cc|c}
\toprule
\multirow{2}{*}{\textbf{Model}} 
& \multicolumn{3}{c|}{\textbf{RefCOCO}} 
& \multicolumn{3}{c|}{\textbf{RefCOCO+}} 
& \multicolumn{2}{c|}{\textbf{RefCOCOg}} 
& \multirow{2}{*}{\textbf{Avg}} \\
\cmidrule(lr){2-4} \cmidrule(lr){5-7} \cmidrule(lr){8-9}
& val & testA & testB 
& val & testA & testB 
& val & test 
&  \\
\midrule

Qwen3-VL-8b-Instruct\tablefootnote{reproduced using lmms-eval~\cite{zhang2024lmmsevalrealitycheckevaluation}}
& \ 91.12 \ & 93.74 \ & 87.52 \ 
& \ 86.78 \ & 90.57 \ & 81.06 \ 
& \ 88.74 \ & 88.73 \ 
& \ 88.70  \\

KARL (mixed)
& \ 92.09 \ & 93.57 \ & 88.60 \ 
& \ 87.47 \ & 91.70 \ & 81.75 \ 
& \ 89.12 \ & 88.89 \ 
& \ 89.28 \\

\bottomrule
\end{tabular}
}
\label{tab:refcoco_res}
\end{table}

\paragraph{Baseline Models.}
When evaluating certain baseline models, some adjustments were made to accommodate minor differences in their output formats.For DeepSeek-VL2 \cite{wu2024deepseek}, the first output bounding box was selected as the model's predicted answer for downstream evaluation. For the evaluation of GroundingDINO-1.6-Pro~\cite{groundingdino_1_6} and DINO-X~\cite{ren2024dinoxunifiedvisionmodel}, we used their official APIs in deepdataspace . Images were converted to base64 format, with oversized images downsampled. Detection targets were fine-grained labels (e.g., “Buick Enclave”) instead of generic categories (e.g., “car”). API requests followed official configurations, and the highest-score bounding box was selected. YOLO-World~\cite{Cheng2024YOLOWorld} was evaluated locally under identical criteria. Regarding Shikra-7B~\cite{chen2023shikra}, we have replicated the evaluation code based on the officially released model and code. However, the actual performance is significantly lower than that reported in the paper. Similar discrepancies have also been observed in replication attempts by other researchers.

\paragraph{Training Details.}

We use Adam optimizer~\cite{adam} with learning rate as $1e-6$, $\beta_1=0.9$ and $\beta_2=0.999$ in the CoT-SFT stage and AdamW optimizer~\cite{loshchilov2018decoupled} with learning rate as $5e-7$, $\beta_1=0.9$ and $\beta_2=0.999$ in the GRPO stage. 
The accumulated batch size is set to 16 in stage 1 and 8 in stage2.
The GRPO stage employs a maximum completion length of 1500 tokens with 4 samples per input.

\subsection{Complementarity with Generic Grounding Data}
We further examine whether our KVG-oriented training can complement, rather than replace, standard grounding supervision.
To this end, we conduct a mixed-training experiment by incorporating a portion of RefCOCO training data into both Stage 1 and Stage 2. Specifically, we include 10,788 RefCOCO training samples in Stage 1 and 2,698 samples in Stage 2.
For the RefCOCO samples, we use direct-answer supervision without CoT rationales in Stage 1, and disable knowledge-aware reward scaling in Stage 2, since these samples do not involve fine-grained entity-level knowledge.

As shown in Tab.~\ref{tab:refcoco_res}, the mixed-training model achieves slightly better average performance on RefCOCO, RefCOCO+, and RefCOCOg than the base Qwen3-VL-8B-Instruct model.
This result suggests that our KVG data and KARL-based optimization are complementary to existing generic grounding supervision: when trained jointly, they can enhance knowledge-intensive grounding ability without sacrificing competitive performance on standard grounding benchmarks.
Notably, our goal here is not to replace generic grounding data with KVG data alone, but to show that the two forms of supervision can be combined effectively within a unified training pipeline.

\subsection{Quantitative Analysis of the Knowledge-Grounding Gap}
\label{sec:appendix-method_gap}
\begin{table}[t]
\caption{Joint correctness statistics of Qwen3-VL-8B-Instruct on the REC-style grounding task and the MCQ-based REG-style knowledge task over KVG-Bench. The relatively large proportion of cases that are correct on REG but incorrect on REC indicates a clear knowledge-grounding gap.}
\label{tab:knowledge_grounding_gap}
\centering
\renewcommand{\arraystretch}{1.2}
\begin{tabular}{lcc}
\toprule
 & \textbf{REC Correct} & \textbf{REC Incorrect} \\
\midrule
\textbf{REG Correct}   & 46.11\% & 22.90\% \\
\textbf{REG Incorrect} & 11.08\% & 19.91\% \\
\bottomrule
\end{tabular}
\end{table}

In the main paper, we qualitatively discuss the \textit{knowledge-grounding gap}, namely that MLLMs may possess the knowledge required to identify a target entity while still failing to ground it correctly.
To further quantify this phenomenon, we compare model performance on two task forms derived from the same KVG-Bench instances: the original \textit{referring expression comprehension} (REC) task for visual grounding, and a multiple-choice \textit{referring expression generation} (REG) variant designed to probe entity-level knowledge.

The REG-style evaluation is constructed following a protocol similar to that used in the knowledge estimation analysis.
Specifically, for each query, we convert the original grounding instance into a multiple-choice question in which the model is asked to identify the entity referred to by the query.
We create three distractor options for each case.
Whenever possible, we first select other entities appearing in the same image as distractors, since these candidates are visually co-present and therefore constitute more challenging alternatives.
If fewer than three such entities are available, we randomly sample the remaining distractors from other entity labels within the same category, ensuring that each question always contains three negative options in total.
This design yields an MCQ-based REG task that preserves the semantic scope of the original KVG query while removing the need for explicit localization.

We adopt this REG-style task as a proxy for knowledge because it mainly evaluates whether the model can correctly identify the target fine-grained entity among semantically related alternatives, without requiring box prediction.
In contrast, REC additionally requires translating such knowledge into precise visual grounding.
Therefore, comparing correctness across REG and REC on the same test cases provides a direct way to examine whether the model's failure stems from lacking the relevant entity knowledge itself, or from failing to effectively use that knowledge during grounding.

As shown in \cref{tab:knowledge_grounding_gap}, only 46.11\% of cases are solved correctly on both tasks, while 19.91\% fail on both.
More importantly, the proportion of cases where the model answers the knowledge task correctly but fails on grounding reaches 22.90\%, which is substantially higher than the opposite case, where grounding is correct but knowledge prediction is wrong (11.08\%).
This asymmetry provides direct quantitative evidence for the knowledge-grounding gap: for a considerable portion of examples, the model already captures the relevant entity knowledge, yet fails to translate such knowledge into accurate visual localization.
In other words, the bottleneck is often not the absence of knowledge itself, but the inability to effectively leverage that knowledge during grounding.

This result also suggests that improving KVG performance requires more than strengthening either visual grounding or factual knowledge in isolation.
A key challenge is to better align knowledge usage with fine-grained grounding decisions, which motivates our emphasis on knowledge-aware reasoning and optimization.


\newpage

\end{document}